\documentclass[11pt,a4paper]{article}
\usepackage[hyperref]{acl2017}
\usepackage{times}
\usepackage{latexsym}
\usepackage{rotating}
\usepackage{graphicx}

\usepackage{url}

\aclfinalcopy
\def\aclpaperid{43}

\newcommand*\rot{\rotatebox{40}}
\newcommand*\rotninety{\rotatebox{90}}

\title{External Evaluation of Event Extraction Classifiers for Automatic Pathway Curation: An extended study of the mTOR pathway}

\author{Wojciech Kusa \\
  AGH University of  \\
  Science and Technology \\
  Cracow, Poland \\
  {\tt wojciechkusa@gmail.com} \\\And
  Michael Spranger \\
  Sony Computer Science  \\
  Laboratories Inc. \\
  Tokyo, Japan \\
  {\tt michael.spranger@gmail.com} \\}

\date{}

\begin{document}
\maketitle
\begin{abstract}
This paper evaluates the impact of various event extraction systems on automatic pathway curation using the popular mTOR pathway. We quantify the impact of training data sets as well as different machine learning classifiers and show that some improve the quality of automatically extracted pathways.
\end{abstract}

\section{Introduction}

Biological pathways encode sequences of biological reactions, such as phosphorylation, activation etc, involving various biological species, such as genes, proteins  \cite{aldridge2006physicochemical, kitano2002systems}. Studying and analyzing pathways is crucial to understanding biological systems and for the development of effective disease treatments and drugs \cite{creixell2015pathway, khatri2012ten}. There have been numerous efforts to reconstruct detailed process-based and disease level pathway maps such as Parkinson disease map (Fujita et al., 2014), Alzheimers disease Map \cite{mizuno2012alzpathway}, mTOR pathway Map \cite{caron2010comprehensive}, and the TLR pathway map \cite{oda2006comprehensive}. Traditionally, these maps are constructed and curated by expert pathway curators who manually read numerous biomedical documents, comprehend and assimilate the knowledge in them and construct the pathway.

With increasing number of scientific publications manual pathway curation is becoming more and more impossible. Therefore, Automated Pathway Curation (APC) and semi-automated biological knowledge extraction has been an active research area \cite{ananiadou2010event,ohta2013overview,szostak2015construction} trying to overcome the limitations of manual curation using various techniques from hand-crafted NLP systems \cite{allen2015complex} to machine learning techniques \cite{bjorne2011extracting}.  Machine-learning NLP systems, in particular, show good performance in BioNLP tasks, but they are still performing less good in automated pathway curation, partly because there have been few attempts to measure the performance of NLP systems for APC directly.

Recently, there has been some attempt at remedying the situation and new datasets and evaluation measures have been proposed. For instance, Spranger et al. \shortcite{spranger2016measuring} use the popular human-generated mTOR pathway map \cite{caron2010comprehensive,efeyan2010mtor,katiyar2009redd1} and quantify the performance of a particular APC system and its ability to recreate the complete pathway automatically. Results reported were mixed.  

One of the key components in such APC systems is identification of triggers, events and their relationships. These machine learning-based systems are essentially just supervised classification components. 

This paper explores whether we can improve results of automated pathway curation for mTOR pathway by using different training datasets and learning algorithms. We show that the choice of event extraction classifiers increases F-score by up to 20\% compared to state-of-the-art system. Our results also show that within limits the choice of training data has significantly less impact on results than the choice of classifier. Our results also suggest that additional research is necessary to solve the problem of APC.

\section{Automatic Pathway Curation}
\label{s:automatic-pathway-extraction}
We constructed an automatic pathway curation system that take as input scientific articles in PDF format and transforms them into SBML encoded, annotated pathway maps. The pipeline has multiple steps.
\begin{enumerate}
\setlength\itemsep{-0.2em}
\item PDFs are translated into pure text files using the cermine\footnote{http://cermine.ceon.pl/index.html} tool. 
\item Preprocessing provides tokenization, POS tagging, dependency and syntax parsing.
\item An \emph{event extraction system} extracts the mentions of entities (genes, proteins etc), reactions (e.g. phosphorylation) and their arguments (theme, cause, product).
\item A converter constructs pathways from the information provided by the event extraction system.
\item An annotation system maps extracted entities and events to Entrez gene identifiers and SBO terms.
\end{enumerate}

The following sections detail steps 3 to 5.

\subsection{Event Extraction}
\label{s:event-extraction}
We used the TURKU Event Extraction System (TEES) for event extraction \citep{bjorne2010complex}. This system is one of the most successful BioNLP systems. It has not only won 1st place in BioNLP competitions but was also the only one NLP system that participated in all BioNLP-ST 2013 tasks \cite{bjorne2012university}. The system combines various NLP techniques to extract information from text. TEES workflow consists of four steps:

\begin{enumerate}
\setlength\itemsep{-0.2em}
\item Trigger Detection - detection of named entities and event triggers in a given sentence to construct nodes of the event graph. 
\item Edge detection - construction of complex events linking few triggers to create event graph. Output produced during this step is often a directed, typed edge connecting two entity nodes.
\item Unmerging - event nodes from merged event graph are duplicated in order to separate arguments into valid combinations. This step is needed for evaluation of final results in BioNLP Shared Task standard.
\item Modifiers detection - final component that defines additional attributes for events such as speculation and negation modifiers.
\end{enumerate}

By default TEES trains a different instance of multiclass Support Vector Machines (SVM) for each step. Recent versions of TEES \cite{bjorne2015tees} allow to easily exchange the SVM classifiers with other supervised classification algorithms. For example, all \textit{scikit-learn} multiclass, supervised learning algorithms that support sparse feature matrices can be applied \citep{scikit-learn}. Thanks to this it is possible to  test different algorithms for event extraction task and automatic pathway extraction. For this paper, we exchanges classifiers in all steps 1-4s as described in Section \ref{par:classifiers}. The output of TEES is a standoff formatted representation of entities and events.

\subsection{Conversion Standoff to SBML pathways}
In principle events and entities extracted by TEES correspond to biological species and reactions. We translate the NLP representation into SBML -- the standard, XML-based markup language for representing biological models \cite{hucka2003systems}. SBML essentially encodes models using biological players called {\tt sbml:species}\footnote{We refer to SBML vocabulary using the prefix ``sbml".}. {\tt sbml:species} can participate in interactions, called {\tt sbml:reaction}.  Species participate in interaction as {\tt sbml:reactant}, {\tt sbml:product} and {\tt sbml:modifier}. The basic idea being that some quantity of reactant is consumed to produce a product. Reactions are influenced by modifiers. The mapping algorithm is adopted from and described in more detail in Spranger et al. \shortcite{spranger2015extracting}. 

\subsection{SBO/GO, Entrez Gene Annotations}
The SBML encoded, automatically extracted pathway is further \emph{annotated} using Systems Biology Ontology (SBO) \citep{le2006model} and Gene Ontology (GO) terms. SBO also provides a class hierarchy for reaction types. For instance, the NLP system identify phosphorylation reactions, which are a subclass of conversion reactions. All reactions in the data are automatically annotated with SBO/GO term (coverage 100\%) using an annotation scheme detailed in \citep{spranger2015extracting}. 

Species (e.g. proteins, genes) were annotated using the gene/protein named entity recognition and normalization software GNAT  \citep{hakenberg2011gnat} - a publicly available gene/protein normalization tool. GNAT returns a set of Entrez Gene identifiers \citep{maglott2005entrez} for each input string. Species were annotated using all returned Entrez Gene identifiers for a particular species (organism human).
We call the set of Entrez Gene identifiers returned by GNAT for each species \emph{Entrez Gene signature}. 

\section{Classifiers for Event Extraction} \label{par:classifiers}
In this paper we evaluate classifiers for event extraction (Section \ref{s:event-extraction}) and their impact on the overall performance of the automatic pathway extraction system. We compare the  following classifiers:

\begin{itemize}
\setlength\itemsep{0em}
\item \textbf{Support Vector Machines (SVM)} is the default TEES classifier \citep{joachims1999making}. It was optimized for linear classification and its performance scales linearly with the number of training examples. 
\item \textbf{Decision Tree (DT)}  creates a model that can predict the target value by learning simple decision rules inferred from the training data. Compared to the other techniques they are relatively fast, cost of using tree is logarithmic in the number of examples. We use Gini impurity criterion to evaluate quality of the split.
\item \textbf{Random Forest (RF)} classifiers fit a number of ensembled decision tree classifiers, each built from a bootstrap sample of a training set. The best split of node is chosen only from a random subset of the features, not all features. Final classifiers are combined by averaging their probabilistic prediction. Single tree have a higher bias but, due to averaging variance of the random forest as a whole decreases.
\item \textbf{Multinomial Naive Bayes (MNNB)} This is an implementation of the naive Bayes algorithm for multinomial data which is one of the classic variants used in classification of discrete features (e.g. text classification). Additive smoothing parameter was set to 1.
\item \textbf{Multi-layer Perceptron (MLP)} MLP is a feedforward neural network model. We use hidden layer with 100 neurons and rectified linear unit activation function. We optimize for logarithmic loss  using stochastic gradient descent. Learning rate is constant and equal to 0.001.
\end{itemize}

For DT, RF, MNNB and MLP we use implementations from \textit{scikit-learn} Python library \citep{scikit-learn}.

\begin{table}[th!]
\begin{center}
\begin{tabular}{l|lll}
Item          & ANN      & GE11  & PC13  \\ \hline
Documents     & 60       & 908   & 260   \\
Words         & 11960    & 205729 & 53811 \\
Entities      & 1921     & 11625 & 7855  \\
Events        & 1284     & 10310 & 5992  \\
Modifiers     & 71       & 1382  & 317   \\
Renaming      & 101      & 571   & 455         
\end{tabular}
\caption{\label{tab:corpora}Corpora statistics}
\begin{tabular}{lrrr}
\hline
{Reaction type}		& 			ANN 		&	  GE11		&  PC13 \\
\hline
Acetylation         &                     0  &         0  &        38  \\
Activation          &                     0  &         0  &       359  \\
Binding             &                   211  &       988  &       606  \\
Catalysis           &                    87  &         0  &         0  \\
Conversion          &                     0  &         0  &       124  \\
Deacetylation       &                     0  &         0  &         1  \\
Degradation         &                     0  &         0  &        49  \\
Demethylation       &                     0  &         0  &         4  \\
Dephosphorylation   &                    14  &         0  &        22  \\
Deubiquitination    &                     0  &         0  &         3  \\
Dissociation        &                    55  &         0  &        54  \\
Gene\_expression     &                    46  &      2265  &       384  \\
Hydroxylation       &                     0  &         0  &         1  \\
Inactivation        &                     0  &         0  &        76  \\
Localization        &                    27  &       281  &        96  \\
Methylation         &                     0  &         0  &         7  \\
Negative\_regulation &                   194  &      1309  &       801  \\
Pathway             &                     0  &         0  &       443  \\
Phosphorylation     &                   252  &       192  &       406  \\
Positive\_regulation &                   235  &      3385  &      1506  \\
Protein\_catabolism  &                    18  &       110  &         0  \\
Regulation          &                   132  &      1113  &       707  \\
Transcription       &                     8  &       667  &        74  \\
Translation         &                     1  &         0  &        11  \\
Transport           &                     0  &         0  &       189  \\
Ubiquitination      &                     4 &         0 &        31 \\
\hline
\end{tabular}
\caption{\label{tab:corpora_reactions}Reaction types annotated for training data sets.}
\end{center}
\end{table}

\section{Datasets}
\subsection{Training Datasets}
In order to quantify the impact of training data, we test the following three training sets.

\begin{itemize}
\setlength\itemsep{0em}
\item \textbf{ANN} - consists of 60 abstracts of scientific papers from Pubmed database related to the mTORpathway map. This dataset was human-annotated for NLP system training \cite[Corpus annotations (c) GENIA Project]{ohta2011pathways} .
\item \textbf{GE11} consists of 908 abstracts and full texts of scientific papers used in BioNLP ST 2011 GENIA Event Extraction task as training data \cite{kim2012genia}. 
\item \textbf{PC13} consists of 260 abstracts of scientific papers used in BioNLP ST 2013 Pathway Curation task as training data \cite{ohta2013overview}. The task goal was to evaluate the applicability of event extraction systems to support the automatic curation and evaluation of biomolecular pathway models.
\end{itemize}

The overall corpora statistics are summarized in Table \ref{tab:corpora}. GE11 and PC13 have the largest number of annotated events. ANN is much smaller in comparison. Also, the distribution of event types differs between data sets (Table \ref{tab:corpora_reactions}). GE11 uses more general terms (Binding, Regulation) compared to PC13 where some specific events appear only a few times (Deacetylation, Hydroxylation, Methylation). 

We train classifiers on four combinations of the three training datasets: 1) standalone GE11; 2) GE11+ANN - combined GE11 and ANN; 3) combined GE11+PC13+ANN - GE11, PC13 and ANN; 4) PC13+ANN - combined PC13 and ANN. For instance, DT+GE11 refers to a decision tree classifier trained on GE11.

We use GE11-Devel BioNLP ST2011 dataset for hyperparameter optimization of all classifiers.

\subsection{Test Data}
Performance of classifiers is tested on the mTOR pathway map \cite{caron2010comprehensive}. The map was constructed by expert human curators using 522 full text papers from the PubMed database. The experts curated a single large map using CellDesigner \cite{funahashi2008celldesigner} - a software for modeling and executing mechanistic models of pathways. CellDesigner represents information using a heavily customized XML-based SBML format  \cite{hucka2003systems}. 

{\bf Target Human expert data} We translate the curator map into standard SBML and further enrich the information using SBO/GO and Entrez Gene annotations. For SBO/GO, we use existing annotations provided by curators and extend them by automatic annotations deduced from reactants and products of reactions. For example,  if a phosphoryl group is added in a reaction, it is annotated using the SBO term for phosphorylation. Each reaction may be annotated with multiple SBO/GO terms. Also we annotate the curated map with Entrez gene identifiers (similar to the automatic extraction data). We call this pathway \emph{TARGET}.

{\bf Testing classifiers} The 522 full text papers -- used by human curators for the construction of the mTOR pathway -- are used for evaluating the different text mining classifiers. For this, we plug in (trained) classifiers into the automatic pathway extraction pipeline which performs preprocessing, event extraction, conversion to SBML and annotation (see  also Section \ref{s:automatic-pathway-extraction}). The output of this is an annotated SBML file that is subsequently compared to human-curated SBML-encoded pathway data.

\section{Evaluation}
Evaluation of the classifiers (and the system as a whole) is performed by comparing the automatically extracted pathway with the hand-curated pathway. Spranger et al. \shortcite{spranger2016measuring} propose a number of graph overlap algorithms for quantifying the difference and similarity of two pathways. Here we employ the same measures. The following summarizes the strategies.

{\bf Species} In order to decide whether species in two pathways are the same, we use the name of the identifiers and their Entrez gene signatures.
\begin{description}\setlength\itemsep{-0.5em}
\item[nmeq: ] Two species are equal if their names are exactly equal. We remove certain prefixes from the names (e.g. phosphorylated).
\item[appeq: ] Two species are equal if their names are approximately equal. Two names are approximately equal iff their Levenshtein-based string distance is above 90 \citep{levenshtein1966binary} 
\item[enteq: ] Two species are equal if their entrez gene identifiers are exactly equal. This basically translates to the two species bqbiol:is identifier sets being exactly the same (order does not matter).
\item[entov: ] Two species are equal if their entrez gene identifiers sets overlap. This basically translates to the two species bqbiol:is identifier sets overlapping.
\item[wc: ] Human curated data contains complex species that contain other species as constituents (species that consist of various proteins etc). \emph{wc} allows species to match with constituents of complexes.
\end{description}

{\bf Reaction} match based on their SBO/GO annotations
\begin{description}
\setlength\itemsep{-0.5em}
\item[sboeq: ] Two reactions are equal iff their signatures are exactly the same. That is, the whole set of SBO/GO terms of one reaction is the same as of the other reaction.
\item[sboov: ] Two reactions are equal, iff their signatures overlap. That is, the intersection of the set of SBO/GO terms of one reaction is with the set of SBO/GO terms of the other reaction is not empty.
\item[sobisa: ] Two reactions are equal, iff there is at least one SBO/GO term in each signature that relate in a is\_a relationship in the SBO reaction type hierarchy. For instance, if there is a phosphorylation reaction and a conversion reaction, then \emph{sboisa} will match because phosphorylation is a subclass of conversion according to the SBO type hierarchy.
\end{description}

{\bf Edges} only match if their labels are strictly equal. So if an edge is a reactant, then it has to be a reactant in the other pathway. Same holds for products and modifiers.

{\bf Subgraph matching strategies} are combinations of matching strategies for species, reactions (and for edges which is always the same). For instance, the matching strategy \emph{nmeq, sboeq} is the most strict and requires that species names are exactly equal and that SBO/GO signatures of reactions are exactly equal. The matching strategy \emph{appeq/enteq/wc, sboisa} is the most loose strategy. In this strategy, two species match if their names are approximately equal or if their Entrez gene identifiers overlap or if any of this applies to one of the constituents of the two species. Two reactions match if any of their SBO/GO terms are in a \emph{is\_a} relationship. We compare a total of 24 matching strategies.

{\bf Subgraph overlap} is computed as follows. For each subgraph in the extracted pathway we search for subgraphs in the human curated data that match according to some subgraph matching strategy.  We use \emph{micro-averaged F-score}, precision and recall \citep{sokolova2009systematic} for quantifying the retrieval results. F-score is used to quantify the overlap of species, reactions and edges. We then macro-average these results to get a \emph{total F-score} quantifying performance of the extraction system as a whole.

\section{Results}
Some classifiers take long to train, so we only have partial results for MLP. However, all other classifiers (DT, MNNB, RF, SVM) finished training on all selected combinations of training data sets.

Since we tested 24 subgraph overlap measures with 18 classifiers, we receive a lot of data that cannot be discussed in detail in this paper. Here, we concentrate on general trends in the data. Code and datasets are published as appropriate\footnote{\url{https://github.com/sbnlp/2017BioNLPEvaluation/}}.

\begin{table*}
\tiny
\centering
\begin{tabular}{lrrrrrrrrr}
\hline
 name               &  \multicolumn{1}{l}{\rot{\# species}} &  \multicolumn{1}{l}{\rot{ \# reactions}} &  \multicolumn{1}{l}{\rot{ \# compartments}} &   \multicolumn{1}{l}{\rot{\# edges }}&   \multicolumn{1}{l}{\rot{\# reactant edges}} &   \multicolumn{1}{l}{\rot{\# edges product}} &   \multicolumn{1}{l}{\rot{\# modifiers}} &   \multicolumn{1}{l}{\rot{\# isolated species}} &   \multicolumn{1}{l}{\rot{\# isolated subgraphs}} \\
\hline
 DT+GE11            &      282361 &         92899 &              201 &    195531 &              89001 &             91895 &              14635 &             118162 &                 187871 \\
 DT+GE11+ANN        &      284187 &         95096 &              188 &    212490 &             100529 &             93886 &              18075 &             115427 &                 184542 \\
 DT+GE11+PC13+ANN   &      289504 &         94496 &              208 &    207447 &              94044 &             93559 &              19844 &             118281 &                 188013 \\
 DT+PC13+ANN        &      279647 &         82977 &               20 &    188325 &              86802 &             82469 &              19054 &             123309 &                 184698 \\
 MLP+GE11+ANN       &      278510 &         88502 &              230 &    193150 &              88655 &             87636 &              16859 &             114541 &                 182456 \\
 MNNB+GE11          &      264413 &         69744 &              202 &    137828 &              61448 &             69250 &               7130 &             139402 &                 198972 \\
 MNNB+GE11+ANN      &      245680 &         45690 &                0 &     86771 &              40102 &             45676 &                993 &             166712 &                 206606 \\
 MNNB+GE11+PC13+ANN &      269008 &         68926 &                0 &    142712 &              70292 &             68894 &               3526 &             151495 &                 203903 \\
 MNNB+PC13+ANN      &      287314 &         76932 &                0 &    183029 &              94693 &             76925 &              11411 &             154210 &                 199844 \\
 RF+GE11            &      227613 &         29573 &                9 &     50444 &              20786 &             29133 &                525 &             178233 &                 206874 \\
 RF+GE11+ANN        &      261414 &         67974 &              347 &    130556 &              57195 &             67271 &               6090 &             136180 &                 199157 \\
 RF+GE11+PC13+ANN   &      203314 &         32075 &                1 &     58083 &              25312 &             31704 &               1067 &             146342 &                 177371 \\
 RF+PC13+ANN        &      236220 &         37018 &                0 &     68559 &              30493 &             36909 &               1157 &             168927 &                 204771 \\
 SVM+GE11           &      288421 &         98938 &              451 &    200595 &              89769 &             97791 &              13035 &             109060 &                 191175 \\
 SVM+GE11+ANN       &      262327 &         81207 &              388 &    169841 &              73033 &             80203 &              16605 &             109862 &                 177023 \\
 SVM+GE11+PC13+ANN  &      275303 &         85435 &              312 &    179661 &              77587 &             84549 &              17525 &             114941 &                 184481 \\
 SVM+PC13+ANN       &      275256 &         82119 &               59 &    177651 &              79239 &             81512 &              16900 &             120729 &                 186122 \\
 TARGET             &        2242 &           777 &                7 &      2457 &               1044 &               892 &                521 &      15 &                 4 \\
\hline
\end{tabular}
\caption{General statistics of all datasets. Number of extracted species, reactions and compartments. Total number of edges and of product, reactant and modifier edges. The table also shows the number of  isolated species and the number of unconnected subgraphs for each pathway. The human curated mTOR pathway \emph{TARGET} numbers are shown in the last row.}
\label{t:general-statistics-dataset}
\end{table*}
\subsection{Extraction Results: Species, Reactions, Subgraphs}
Generally speaking the extracted pathways contain two order of magnitudes more species reactions, and edges than the \emph{TARGET} pathway (see Table \ref{t:general-statistics-dataset} for all results). This is normal since the extracted pathways consist of all combinations of entity and event mentions in text. The same entities may occur more often in the text then they are referenced in the actual pathway. 

Our results show that extraction classifiers perform inconsistent with respect to the identification of compartments. While some classifiers retrieve a lot of compartment information (via localization events), others (especially MNNB trained on ANN and PC13 datasets) do not extract any compartments. MNNB with our parameter choice might not be able to learn many different event types so it skips least frequent reaction types (one of which is localization event).

Measuring how many subgraphs there are per pathway, we can see that more than half of all species extracted by classifiers are isolated and not connected to any reactions. Similarly we see many (small) subgraphs being extracted by the classifiers, whereas \emph{TARGET} consists of essentially one large connected graph (with a few modeling mistakes). 

\begin{table}[ht!]
\begin{tabular}{lrr}
\hline
                        &  this & Spr16 \\
 &  f-score & f-score \\
\hline
 nmeq, sboeq            &          {\bf 11.7} &                    7.6 \\
 nmeq, sboov            &          {\bf 15.3} &                   11.4 \\
 nmeq, sboisa           &          {\bf 18.1} &                   13.6 \\
 appeq, sboeq           &          {\bf 12.5} &                    8.1 \\
 appeq, sboov           &          {\bf 16.3} &                   12.0 \\
 appeq, sboisa          &          {\bf 19.4} &                   14.5 \\
 appeq/enteq, sboeq     &          {\bf 16.9} &                   11.9 \\
 appeq/enteq, sboov     &          {\bf 21.7} &                   17.1 \\
 appeq/enteq, sboisa    &          {\bf 26.0} &                   20.4 \\
 appeq/entov, sboeq     &          {\bf 36.2} &                   26.9 \\
 appeq/entov, sboov     &          {\bf 41.9} &                   34.7 \\
 appeq/entov, sboisa    &          {\bf 48.6} &                   39.5 \\
 nmeq/wc, sboeq         &          {\bf 23.3} &                   15.0 \\
 nmeq/wc, sboov         &          {\bf 26.0} &                   19.6 \\
 nmeq/wc, sboisa        &          {\bf 29.1} &                   22.0 \\
 appeq/wc, sboeq        &          {\bf 24.6} &                   15.7 \\
 appeq/wc, sboov        &          {\bf 27.4} &                   20.4 \\
 appeq/wc, sboisa       &          {\bf 30.9} &                   23.1 \\
 appeq/enteq/wc, sboeq  &          {\bf 39.7} &                   29.1 \\
 appeq/enteq/wc, sboov  &          {\bf 45.3} &                   37.2 \\
 appeq/enteq/wc, sboisa &          {\bf 52.0} &                   42.2 \\
 appeq/entov/wc, sboeq  &          {\bf 39.7} &                   29.1 \\
 appeq/entov/wc, sboov  &          {\bf 45.3} &                   37.2 \\
 appeq/entov/wc, sboisa &          {\bf 52.0} &                   42.2 \\
 \hline
 \end{tabular}
\caption{This table compares macro F-score performance of the classifiers discussed in this paper with results reported in Spranger et al. \shortcite{spranger2016measuring}}
\label{f:comparison-last-year}
\end{table}

\subsection{General Trends Subgraphs overlap} 

\begin{figure}[ht!]
\centering
\includegraphics[width=1\columnwidth]{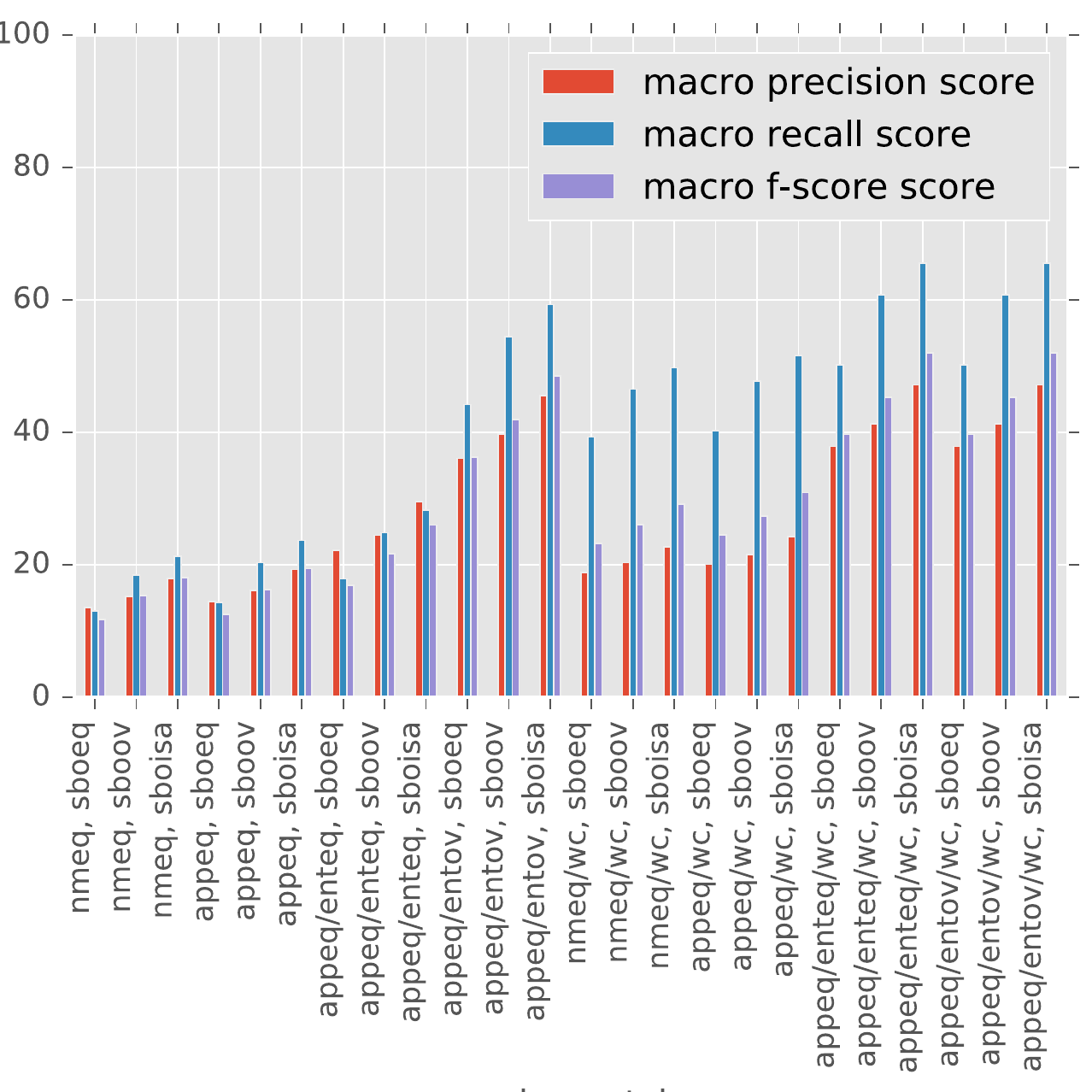}
\caption{Best performing classifier f-score, precision and recall for each subgraph overlap functions. The x-axis are the different subgraph overlap function. The y-axis shows precision, recall, f-score of the highest classifier for each subgraph overlap function. Notice that these can be different classifiers for each subgraph overlap function (see Table \ref{t:macro-f-score-results} for all results).}
\label{f:best-node-match-performance}
\end{figure}

Let us first concentrate on overall performance especially with respect to previous results. For this we compute the best classifiers and their score for different matching strategies. For each matching strategy, we evaluate all classifiers and then choose the best performing one and compare it with the results reported in Spranger et al. \shortcite{spranger2016measuring}/Spr16. Table \ref{f:comparison-last-year} shows that the best classifiers outperform Spr16 in all cases and for some subgraph overlap measures by 10 points.

If we analyze the classifiers from this paper in more detail, results (Figure \ref{f:best-node-match-performance}, Table \ref{t:macro-f-score-results}) show that for the strictest matching strategy (\emph{nmeq, sboeq}) the best classifiers reach a macro F-score of 12 (with 14 precision, 13 recall scores). For the loosest strategy (\emph{appeq/entov/wc, sboisa}) this goes up to F-score 52 (47 precision, 66 recall). These results show that when it comes to \emph{exact} extraction the classifiers fail badly, whereas with more looser overlap strategies, performance becomes reasonable and there is some overlap between the extracted and the human-curated data. Of course, this also entails that the automatically extracted pathway does not completely capture what humans are constructing from the text.

Generally speaking overlap strategies that are loose with respect to constituents of complex species (\emph{wc}) outperform their non \emph{wc} counterparts. For instance, \emph{nmeq/wc, sboeq} performs much better than \emph{nmeq, sboeq}. This shows that complex species are important for the mTOR pathway but their extraction is not very detailed - which is why the overlap matching strategy has to be lenient with respect to complex species constituents. The increase in F-score for \emph{wc} matching strategies is primarily driven by an increase in recall score. For instance, the difference between \emph{nmeq, sboeq} and \emph{nmeq/wc, sboeq} is more than 20 points, whereas precision does not improve that much. The reasons for that is that the same subgraphs in the extracted pathway overlap with more subgraphs in \emph{TARGET}. So it is not the case that other subgraphs in the extracted pathway overlap with \emph{TARGET}. 

Results also show that recall is in general much higher than precision for looser strategies. For instance, \emph{wc} strategies (right hand side of Figure \ref{f:best-node-match-performance}) double the recall score w.r.t to their precision scores. This also shows that in principle loosening matching strategies impacts mostly recall as the same subgraphs in the extracted data overlap with the human curated data.

\begin{figure}[ht!]
\centering
\includegraphics[width=1\columnwidth]{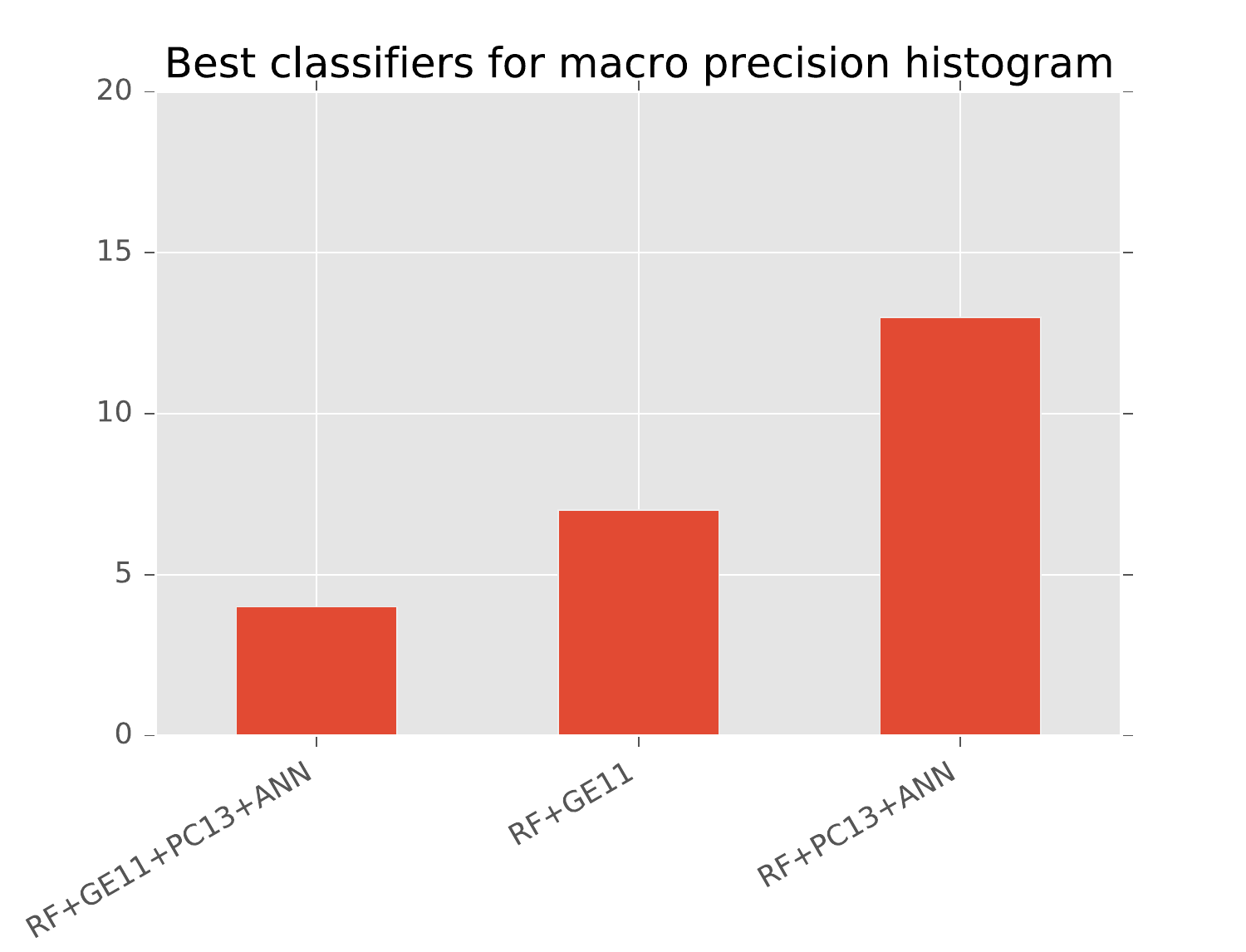}
\includegraphics[width=1\columnwidth]{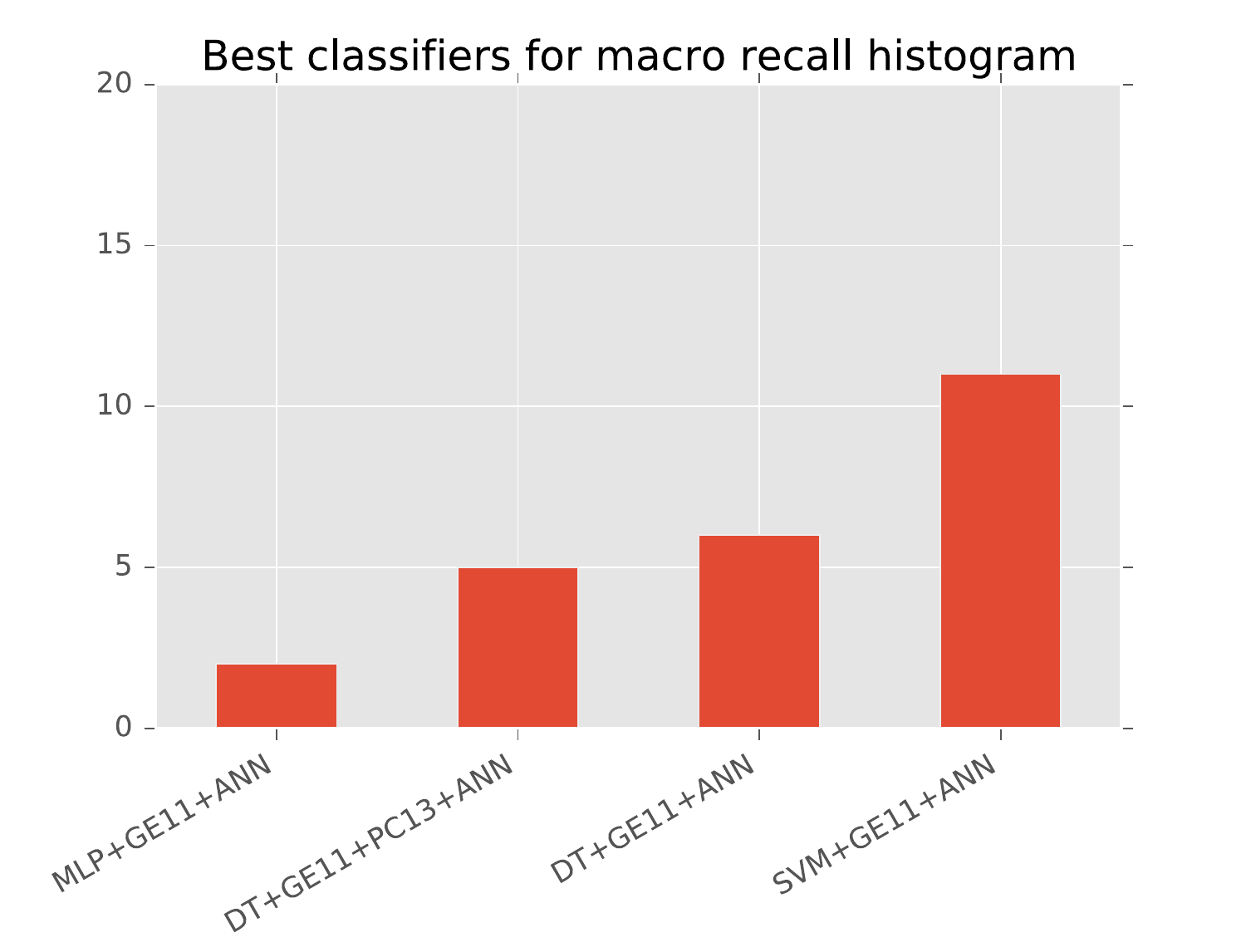}
\includegraphics[width=1\columnwidth]{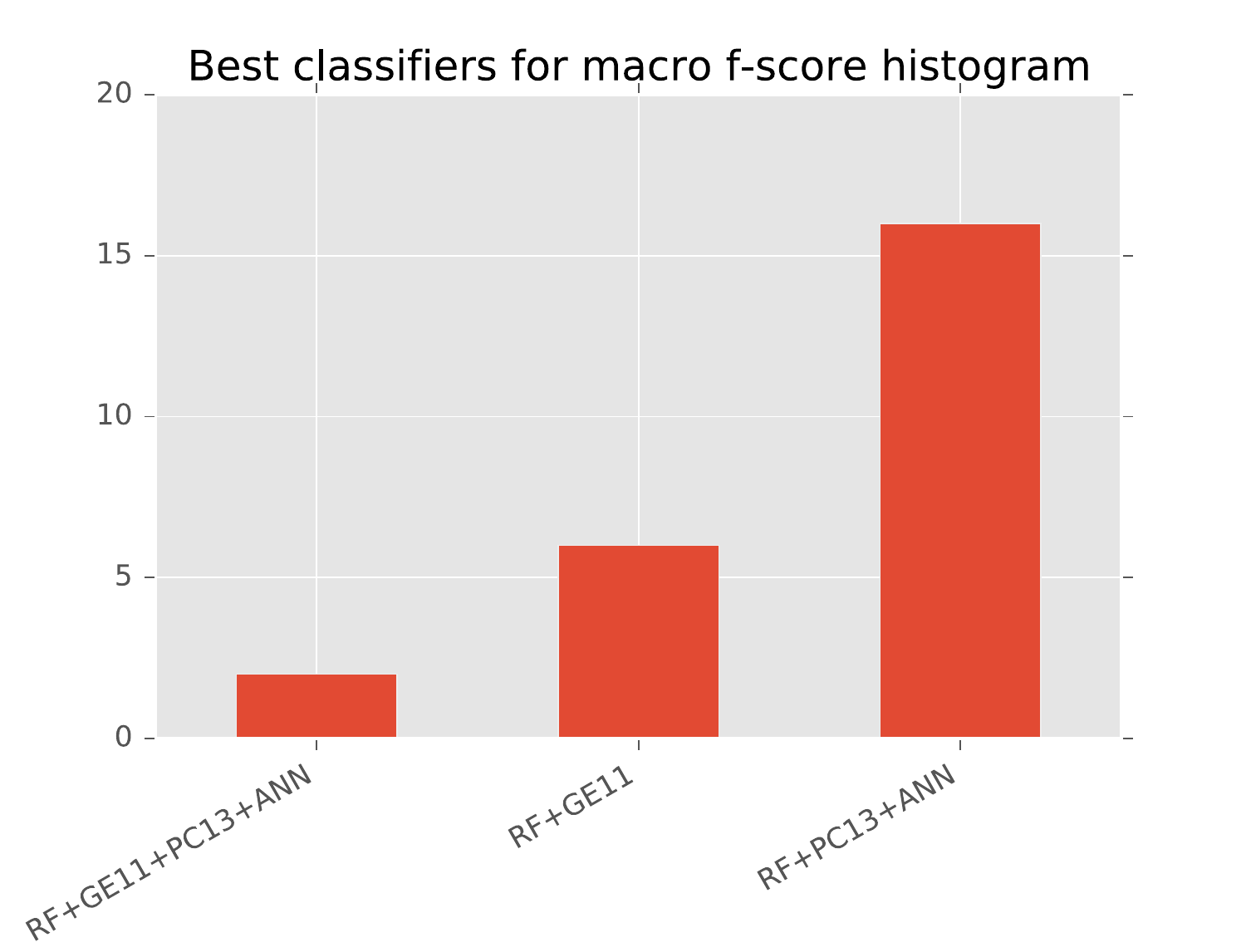}
\caption{Histogram of best classifiers. This histogram is generated by counting how often a classifier is the best for a particular subgraph matching strategy.}
\label{f:best-classifiers}
\end{figure}

\subsection{Classifier Performance in Detail} 
The bottom figure in Figure \ref{f:best-classifiers} shows the best classifiers in terms of precision, recall and F-score. We measured how often a classifier is the best classifier (for each of the 24 subgraph overlap strategies). It is clear that overall Random Forest classifier (RF) performance is the best. For all 24 matching strategies it is a Random Forest classifier that is better than any other competitor with RF trained on PC13 and ANN being the most frequent best classifier overall. Second place is Random Forest trained simply on GE11 (the largest dataset in terms of entities and events). No other classifiers (SVM, MLP, MNNB, DT) outperform RF. Training on all datasets (RF+GE11+PC13+ANN) does not seem to increase success significantly. Performance across different RF classifiers is on par and good (see Table \ref{t:macro-f-score-results})

Results in the top figure of Figure \ref{f:best-classifiers} show that RF has the best precision performance. RF+PC13+ANN is the most frequent best classifier w.r.t precision. RF+GE11 and RF+GE11+PC13+ANN also performing comparably. Compared to recall this means that RF wins F-score because they are best in precision.

No RF classifier performs best in recall. Results show that MLP, DT and SVM all perform well for certain subgraph overlap strategies with SVM being most often the best classifier, followed by various DT-based classifiers and MLP.

\begin{figure}
\centering
\includegraphics[width=.835\columnwidth]{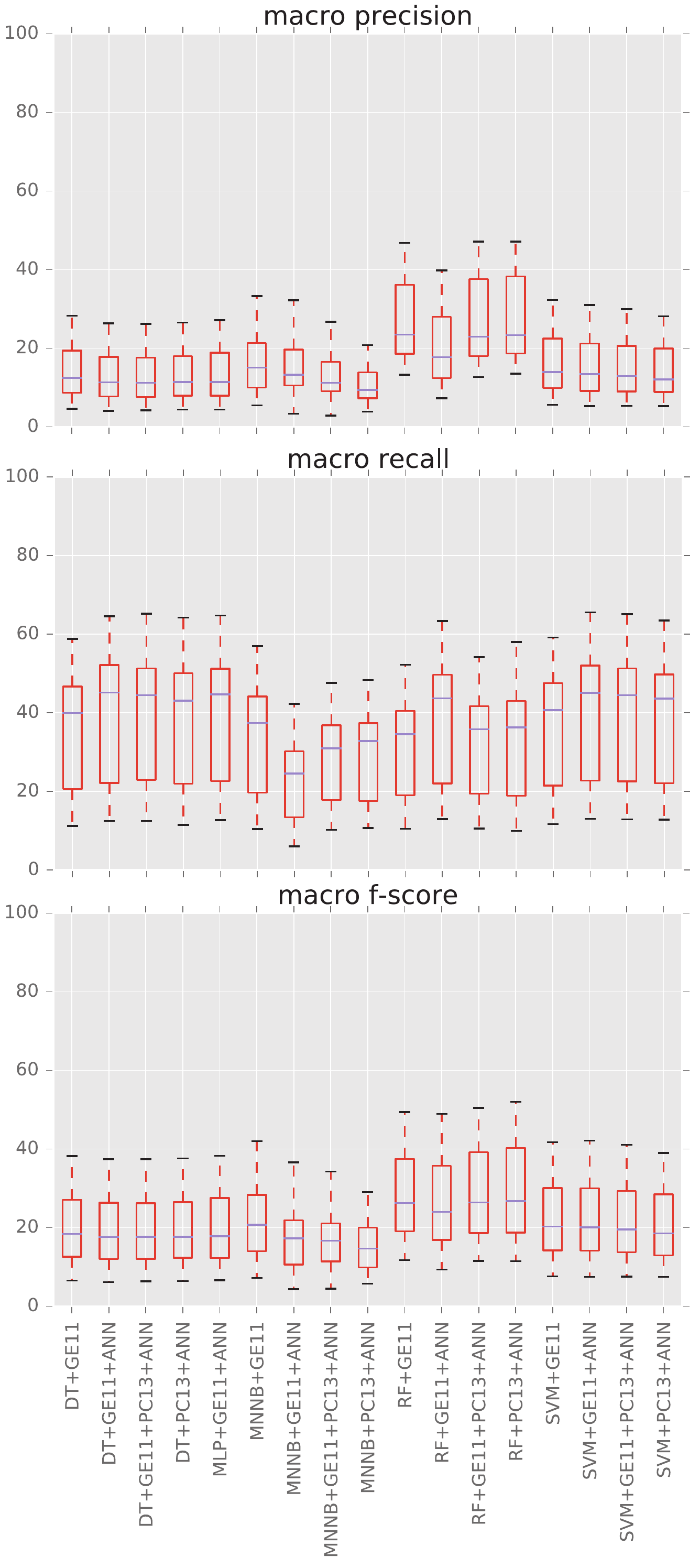}
\caption{Statistics of classifier performance across all matching strategies. X-axis - classifiers. Y-axis - macro precision top, macro recall middle and macro f-score bottom (with 100 being perfect score).}
\label{f:box-performance}
\end{figure}

Figure \ref{f:box-performance} gives results for all classifiers across all matching strategies. Looser strategies give the max and strict matching strategies the min data points. We can see that performance is primarily driven by the choice of classifier as the F-score mostly varies with the type of classifier used (even though there are a few outliers). Situation is a bit more varied for precision and recall. Interestingly choice of dataset seems to have less impact. Generally speaking MNNB are the least successful. RF clearly dominate precision on average but are close enough to DT and SVM on recall.

\section{Conclusion}
This paper continues the current trend of extending NLP systems for APC and building more complete systems that allow evaluation with respect to some external standard - here the hand curated mTOR pathway. 

We measured the impact of different classifiers on retrieval performance and showed that certain classifiers have the potential to increase retrieval performance. Especially Random Forest classifiers perform much better on mTOR than previously tried Support Vector Machines. On the other hand, the training data choice seems to have little impact (at least for the tested ANN, GE11 and PC13 training datasets).

Spranger et al. \shortcite{spranger2016measuring} argue that not all of the problems of APC can be overcome by using more training data on event extraction systems. They argue that additions such as complex species recognition, co-reference resolution and pathway construction are needed to ultimately solve the problem posed by APC. This certainly remains true and is not directly questioned by results in this paper. The system described here does not automatically compose single pathway maps from the extracted data. Nevertheless, our results suggest that a lot of progress can be made by improving on the event extraction part of the pipeline.

This paper focuses on evaluating current machine learning techniques for event extraction. We are currently in the process of evaluating other systems including rule-based ones.

\section*{Acknowledgments}
We would like to thank the authors of the Turku Event Extraction System for providing such an excellent system to the community. We also thank SBI for making the mTOR dataset available and their collaboration on evaluation and measurements.

\bibliography{acl2017}

\begin{thebibliography}{}
\expandafter\ifx\csname natexlab\endcsname\relax\def\natexlab#1{#1}\fi

\bibitem[{Aldridge et~al.(2006)Aldridge, Burke, Lauffenburger, and
  Sorger}]{aldridge2006physicochemical}
B.~B Aldridge, J.~M Burke, D.~A Lauffenburger, and P.~K Sorger. 2006.
\newblock Physicochemical modelling of cell signalling pathways.
\newblock {\em Nature cell biology\/} 8(11):1195--1203.

\bibitem[{Allen et~al.(2015)Allen, de~Beaumont, Galescu, and
  Teng}]{allen2015complex}
J.~Allen, W.~de~Beaumont, L.~Galescu, and C.~M. Teng. 2015.
\newblock Complex event extraction using drum.
\newblock {\em BioNLP 2015 Workshop on Biomedical Natural Language
  Processing\/} pages 1--11.

\bibitem[{Ananiadou et~al.(2010)Ananiadou, Pyysalo, Tsujii, and
  Kell}]{ananiadou2010event}
S.~Ananiadou, S.~Pyysalo, J.~Tsujii, and D.~Kell. 2010.
\newblock Event extraction for systems biology by text mining the literature.
\newblock {\em Trends in biotechnology\/} 28(7):381--90.

\bibitem[{Bj{\"o}rne et~al.(2010)Bj{\"o}rne, Ginter, Pyysalo, Tsujii, and
  Salakoski}]{bjorne2010complex}
J.~Bj{\"o}rne, F.~Ginter, S.~Pyysalo, J.~Tsujii, and T.~Salakoski. 2010.
\newblock Complex event extraction at pubmed scale.
\newblock {\em Bioinformatics\/} 26(12):i382--i390.

\bibitem[{Bj{\"o}rne et~al.(2012)Bj{\"o}rne, Ginter, and
  Salakoski}]{bjorne2012university}
J.~Bj{\"o}rne, F.~Ginter, and T.~Salakoski. 2012.
\newblock University of turku in the bionlp'11 shared task.
\newblock {\em BMC bioinformatics\/} 13(11):1.

\bibitem[{Bj{\"o}rne et~al.(2011)Bj{\"o}rne, Heimonen, Ginter, Airola,
  Pahikkala, and Salakoski}]{bjorne2011extracting}
J.~Bj{\"o}rne, J.~Heimonen, F.~Ginter, A.~Airola, T.~Pahikkala, and
  T.~Salakoski. 2011.
\newblock Extracting contextualized complex biological events with rich
  graph-based feature sets.
\newblock {\em Computational Intelligence\/} 27(4):541--557.

\bibitem[{Bj{\"o}rne and Salakoski(2015)}]{bjorne2015tees}
J.~Bj{\"o}rne and T.~Salakoski. 2015.
\newblock Tees 2.2: Biomedical event extraction for diverse corpora.
\newblock {\em BMC Bioinformatics\/} 16(16):S4.

\bibitem[{Caron et~al.(2010)Caron, Ghosh, Matsuoka
  et~al.}]{caron2010comprehensive}
E.~Caron, S.~Ghosh, Y.~Matsuoka, et~al. 2010.
\newblock A comprehensive map of the mtor signaling network.
\newblock {\em Molecular systems biology\/} 6(1).

\bibitem[{Creixell et~al.(2015)Creixell, Reimand, Haider
  et~al.}]{creixell2015pathway}
P.~Creixell, J.~Reimand, S.~Haider, et~al. 2015.
\newblock Pathway and network analysis of cancer genomes.
\newblock {\em Nature methods\/} 12(7):615.

\bibitem[{Efeyan and Sabatini(2010)}]{efeyan2010mtor}
A.~Efeyan and D.~Sabatini. 2010.
\newblock mtor and cancer: many loops in one pathway.
\newblock {\em Current opinion in cell biology\/} 22(2):169--176.

\bibitem[{Funahashi et~al.(2008)Funahashi, Matsuoka, Jouraku
  et~al.}]{funahashi2008celldesigner}
A.~Funahashi, Y.~Matsuoka, A.~Jouraku, et~al. 2008.
\newblock Celldesigner 3.5: a versatile modeling tool for biochemical networks.
\newblock {\em Proceedings of the IEEE\/} 96(8):1254--1265.

\bibitem[{Hakenberg et~al.(2011)Hakenberg, Gerner, Haeussler, Solt, Plake,
  Schroeder, Gonzalez, Nenadic, and Bergman}]{hakenberg2011gnat}
J.~Hakenberg, M.~Gerner, M.~Haeussler, I.~Solt, C.~Plake, M.~Schroeder,
  G.~Gonzalez, G.~Nenadic, and C.~M Bergman. 2011.
\newblock The gnat library for local and remote gene mention normalization.
\newblock {\em Bioinformatics\/} 27(19):2769--2771.

\bibitem[{Hucka et~al.(2003)Hucka, Finney, Sauro et~al.}]{hucka2003systems}
M.~Hucka, A.~Finney, H.~Sauro, et~al. 2003.
\newblock The systems biology markup language (sbml): a medium for
  representation and exchange of biochemical network models.
\newblock {\em Bioinformatics\/} 19(4):524--531.

\bibitem[{Joachims(1999)}]{joachims1999making}
T.~Joachims. 1999.
\newblock Making large-scale {SVM} learning practical.
\newblock In B.~Sch{\"o}lkopf, C.~Burges, and A.~Smola, editors, {\em Advances
  in Kernel Methods - Support Vector Learning\/}, MIT Press, Cambridge, MA,
  chapter~11, pages 169--184.

\bibitem[{Katiyar et~al.(2009)Katiyar, Liu, Knutzen et~al.}]{katiyar2009redd1}
S.~Katiyar, E.~Liu, C.~Knutzen, et~al. 2009.
\newblock Redd1, an inhibitor of mtor signalling, is regulated by the
  cul4a--ddb1 ubiquitin ligase.
\newblock {\em EMBO reports\/} 10(8):866--872.

\bibitem[{Khatri et~al.(2012)Khatri, Sirota, and Butte}]{khatri2012ten}
P.~Khatri, M.~Sirota, and A.~Butte. 2012.
\newblock Ten years of pathway analysis: current approaches and outstanding
  challenges.
\newblock {\em PLoS Comput Biol\/} 8(2).

\bibitem[{Kim et~al.(2012)Kim, Nguyen, Wang, Tsujii, Takagi, and
  Yonezawa}]{kim2012genia}
J.-D. Kim, N.~Nguyen, Y.~Wang, J.~Tsujii, T.~Takagi, and A.~Yonezawa. 2012.
\newblock The genia event and protein coreference tasks of the bionlp shared
  task 2011.
\newblock {\em BMC Bioinformatics\/} 13(11):S1.

\bibitem[{Kitano(2002)}]{kitano2002systems}
H.~Kitano. 2002.
\newblock Systems biology: a brief overview.
\newblock {\em Science\/} 295(5560):1662--1664.

\bibitem[{Le~Nov{\`e}re(2006)}]{le2006model}
N.~Le~Nov{\`e}re. 2006.
\newblock Model storage, exchange and integration.
\newblock {\em BMC neuroscience\/} .

\bibitem[{Levenshtein(1966)}]{levenshtein1966binary}
V.~I Levenshtein. 1966.
\newblock Binary codes capable of correcting deletions, insertions, and
  reversals.
\newblock In {\em Soviet physics doklady\/}. volume~10, pages 707--710.

\bibitem[{Maglott et~al.(2005)Maglott, Ostell, Pruitt, and
  Tatusova}]{maglott2005entrez}
D.~Maglott, J.~Ostell, K.~D Pruitt, and T.~Tatusova. 2005.
\newblock Entrez gene: gene-centered information at ncbi.
\newblock {\em Nucleic acids research\/} 33(suppl 1):D54--D58.

\bibitem[{Mizuno et~al.(2012)Mizuno, Iijima, Ogishima
  et~al.}]{mizuno2012alzpathway}
S.~Mizuno, R.~Iijima, S.~Ogishima, et~al. 2012.
\newblock Alzpathway: a comprehensive map of signaling pathways of alzheimer's
  disease.
\newblock {\em BMC systems biology\/} 6(1):52.

\bibitem[{Oda and Kitano(2006)}]{oda2006comprehensive}
K.~Oda and H.~Kitano. 2006.
\newblock A comprehensive map of the toll-like receptor signaling network.
\newblock {\em Molecular systems biology\/} 2(1).

\bibitem[{Ohta et~al.(2013)Ohta, Pyysalo, Rak et~al.}]{ohta2013overview}
T.~Ohta, S.~Pyysalo, R.~Rak, et~al. 2013.
\newblock Overview of the pathway curation (pc) task of bionlp shared task
  2013.
\newblock In {\em Proceedings of the BioNLP Shared Task 2013 Workshop\/}. ACL,
  pages 67--75.

\bibitem[{Ohta et~al.(2011)Ohta, Pyysalo, and Tsujii}]{ohta2011pathways}
T.~Ohta, S.~Pyysalo, and J.~Tsujii. 2011.
\newblock From pathways to biomolecular events: opportunities and challenges.
\newblock In {\em Proceedings of BioNLP 2011 Workshop\/}. ACL, pages 105--113.

\bibitem[{Pedregosa et~al.(2011)Pedregosa, Varoquaux, Gramfort, Michel,
  Thirion, Grisel, Blondel, Prettenhofer, Weiss, Dubourg, Vanderplas, Passos,
  Cournapeau, Brucher, Perrot, and Duchesnay}]{scikit-learn}
F.~Pedregosa, G.~Varoquaux, A.~Gramfort, V.~Michel, B.~Thirion, O.~Grisel,
  M.~Blondel, P.~Prettenhofer, R.~Weiss, V.~Dubourg, J.~Vanderplas, A.~Passos,
  D.~Cournapeau, M.~Brucher, M.~Perrot, and E.~Duchesnay. 2011.
\newblock Scikit-learn: Machine learning in {P}ython.
\newblock {\em Journal of Machine Learning Research\/} 12:2825--2830.

\bibitem[{Sokolova and Lapalme(2009)}]{sokolova2009systematic}
M.~Sokolova and G.~Lapalme. 2009.
\newblock A systematic analysis of performance measures for classification
  tasks.
\newblock {\em Information Processing \& Management\/} 45(4):427--437.

\bibitem[{Spranger et~al.(2015)Spranger, Palaniappan, and
  Ghosh}]{spranger2015extracting}
M.~Spranger, S.~Palaniappan, and S.~Ghosh. 2015.
\newblock Extracting biological pathway models from nlp event representations.
\newblock In {\em Proceedings of the 2015 Workshop on Biomedical Natural
  Language Processing (BioNLP 2015)\/}. ACL, pages 42---51.

\bibitem[{{Spranger} et~al.(2016){Spranger}, {Palaniappan}, and
  {Ghosh}}]{spranger2016measuring}
M.~{Spranger}, S.~K. {Palaniappan}, and S.~{Ghosh}. 2016.
\newblock {Measuring the State of the Art of Automated Pathway Curation Using
  Graph Algorithms - A Case Study of the mTOR Pathway}.
\newblock {\em ArXiv e-prints\/} .

\bibitem[{Szostak et~al.(2015)Szostak, Ansari, Madan, Fluck, Talikka, Iskandar,
  De~Leon, Hofmann-Apitius, Peitsch, and Hoeng}]{szostak2015construction}
J.~Szostak, S.~Ansari, S.~Madan, J.~Fluck, M.~Talikka, A.~Iskandar, H.~De~Leon,
  M.~Hofmann-Apitius, M.~C Peitsch, and J.~Hoeng. 2015.
\newblock Construction of biological networks from unstructured information
  based on a semi-automated curation workflow.
\newblock {\em Database\/} 2015:bav057.

\end{thebibliography}
\bibliographystyle{acl_natbib}

\begin{sidewaystable*}
\begin{tabular}{l|r|r|r|r|r|r|r|r|r|r|r|r|r|r|r|r|r|}
\hline
                        &   \multicolumn{1}{l|}{\rotninety{DT+GE11}} &   \multicolumn{1}{l|}{\rotninety{DT+GE11+ANN}} &   \multicolumn{1}{l|}{\rotninety{DT+GE11+PC13+ANN}} &   \multicolumn{1}{l|}{\rotninety{DT+PC13+ANN}} &   \multicolumn{1}{l|}{\rotninety{MLP+GE11+ANN}} &   \multicolumn{1}{l|}{\rotninety{MNNB+GE11}} &   \multicolumn{1}{l|}{\rotninety{MNNB+GE11+ANN}} &   \multicolumn{1}{l|}{\rotninety{MNNB+GE11+PC13+ANN}} &   \multicolumn{1}{l|}{\rotninety{MNNB+PC13+ANN}} &   \multicolumn{1}{l|}{\rotninety{RF+GE11}} &  \multicolumn{1}{l|}{\rotninety{RF+GE11+ANN}} &   \multicolumn{1}{l|}{\rotninety{RF+GE11+PC13+ANN}} &   \multicolumn{1}{l|}{\rotninety{RF+PC13+ANN}} &   \multicolumn{1}{l|}{\rotninety{SVM+GE11}} &   \multicolumn{1}{l|}{\rotninety{SVM+GE11+ANN}} &   \multicolumn{1}{l|}{\rotninety{SVM+GE11+PC13+ANN}} &   \multicolumn{1}{l|}{\rotninety{SVM+PC13+ANN}} \\
\hline
nmeq, sboeq            & 6.5                          & 6.1                              & 6.3                                   & 6.4                              & 6.6                               & 7.2                            & 4.3                                & 4.5                                     & 5.7                                & \textbf{11.7}                & 9.3                              & 11.5                                  & 11.5                             & 7.6                           & 7.5                               & 7.5                                    & 7.5                               \\
nmeq, sboov            & 10.1                         & 9.5                              & 9.8                                   & 9.6                              & 9.7                               & 12.0                           & 10.1                               & 11.0                                    & 9.5                                & \textbf{15.3}                & 13.8                             & 15.1                                  & 14.6                             & 11.4                          & 11.4                              & 11.1                                   & 10.2                              \\
nmeq, sboisa           & 11.9                         & 11.4                             & 11.7                                  & 11.7                             & 11.6                              & 14.2                           & 12.2                               & 12.3                                    & 10.2                               & \textbf{18.1}                & 16.1                             & 17.6                                  & 17.7                             & 13.5                          & 13.5                              & 13.2                                   & 12.3                              \\
appeq, sboeq           & 7.0                          & 6.6                              & 6.8                                   & 7.0                              & 7.1                               & 7.5                            & 4.5                                & 4.6                                     & 6.0                                & \textbf{12.5}                & 10.0                             & 12.4                                  & 12.4                             & 8.0                           & 7.9                               & 8.0                                    & 8.0                               \\
appeq, sboov           & 10.8                         & 10.1                             & 10.4                                  & 10.4                             & 10.3                              & 12.5                           & 10.8                               & 11.5                                    & 10.0                               & \textbf{16.3}                & 14.5                             & 16.1                                  & 15.7                             & 12.0                          & 12.0                              & 11.7                                   & 10.8                              \\
appeq, sboisa          & 12.8                         & 12.2                             & 12.5                                  & 12.6                             & 12.5                              & 14.9                           & 13.1                               & 13.0                                    & 10.7                               & \textbf{19.4}                & 17.1                             & 18.9                                  & 19.1                             & 14.4                          & 14.4                              & 14.1                                   & 13.1                              \\
appeq/enteq, sboeq     & 10.2                         & 9.9                              & 10.2                                  & 10.4                             & 10.5                              & 10.6                           & 6.0                                & 6.0                                     & 8.3                                & 16.8                         & 14.1                             & \textbf{16.9}                         & 16.7                             & 11.7                          & 11.5                              & 11.7                                   & 11.8                              \\
appeq/enteq, sboov     & 15.2                         & 14.7                             & 15.0                                  & 14.9                             & 15.0                              & 17.2                           & 14.1                               & 14.9                                    & 13.6                               & 21.6                         & 20.0                             & \textbf{21.7}                         & 21.2                             & 17.0                          & 16.9                              & 16.6                                   & 15.7                              \\
appeq/enteq, sboisa    & 18.1                         & 17.6                             & 18.0                                  & 18.0                             & 17.9                              & 20.6                           & 17.4                               & 17.1                                    & 14.7                               & 25.9                         & 23.5                             & 25.4                                  & \textbf{26.0}                    & 20.2                          & 20.1                              & 19.8                                   & 18.9                              \\
appeq/entov, sboeq     & 23.6                         & 23.0                             & 22.9                                  & 23.3                             & 24.2                              & 24.1                           & 17.2                               & 16.3                                    & 17.2                               & 33.8                         & 31.6                             & 35.2                                  & \textbf{36.2}                    & 26.4                          & 26.2                              & 25.7                                   & 25.3                              \\
appeq/entov, sboov     & 31.0                         & 30.1                             & 30.2                                  & 30.2                             & 31.0                              & 34.2                           & 29.6                               & 29.2                                    & 25.9                               & 39.9                         & 40.6                             & 41.6                                  & \textbf{41.9}                    & 34.1                          & 34.2                              & 33.2                                   & 31.3                              \\
appeq/entov, sboisa    & 35.5                         & 34.6                             & 34.9                                  & 35.0                             & 35.6                              & 39.0                           & 33.9                               & 32.3                                    & 27.4                               & 46.4                         & 45.9                             & 47.2                                  & \textbf{48.6}                    & 38.9                          & 39.2                              & 38.2                                   & 36.4                              \\
nmeq/wc, sboeq         & 13.4                         & 12.4                             & 12.2                                  & 12.5                             & 12.9                              & 13.6                           & 6.9                                & 6.4                                     & 8.4                                & 22.0                         & 17.9                             & 22.3                                  & \textbf{23.3}                    & 14.9                          & 14.4                              & 13.9                                   & 14.0                              \\
nmeq/wc, sboov         & 17.9                         & 16.9                             & 16.6                                  & 16.5                             & 16.9                              & 20.1                           & 16.8                               & 16.4                                    & 14.1                               & 25.4                         & 23.5                             & 25.8                                  & \textbf{26.0}                    & 19.5                          & 19.3                              & 18.3                                   & 17.2                              \\
nmeq/wc, sboisa        & 20.1                         & 19.1                             & 18.8                                  & 19.0                             & 19.1                              & 22.5                           & 18.5                               & 17.6                                    & 14.7                               & 28.5                         & 26.2                             & 28.5                                  & \textbf{29.1}                    & 22.0                          & 21.8                              & 20.8                                   & 19.7                              \\
appeq/wc, sboeq        & 14.1                         & 13.1                             & 12.9                                  & 13.2                             & 13.7                              & 14.2                           & 7.1                                & 6.6                                     & 8.8                                & 23.2                         & 18.9                             & 23.4                                  & \textbf{24.6}                    & 15.6                          & 15.0                              & 14.7                                   & 14.8                              \\
appeq/wc, sboov        & 18.7                         & 17.6                             & 17.3                                  & 17.3                             & 17.7                              & 20.8                           & 17.3                               & 16.9                                    & 14.6                               & 26.7                         & 24.5                             & 27.0                                  & \textbf{27.4}                    & 20.3                          & 20.0                              & 19.2                                   & 18.1                              \\
appeq/wc, sboisa       & 21.1                         & 20.1                             & 19.8                                  & 20.1                             & 20.2                              & 23.5                           & 19.3                               & 18.3                                    & 15.2                               & 30.3                         & 27.5                             & 30.1                                  & \textbf{30.9}                    & 23.1                          & 22.8                              & 22.0                                   & 20.8                              \\
appeq/enteq/wc, sboeq  & 25.7                         & 25.1                             & 24.8                                  & 25.3                             & 26.3                              & 26.4                           & 18.3                               & 16.9                                    & 18.0                               & 36.7                         & 34.1                             & 38.3                                  & \textbf{39.7}                    & 28.7                          & 28.6                              & 28.0                                   & 27.5                              \\
appeq/enteq/wc, sboov  & 33.5                         & 32.7                             & 32.5                                  & 32.4                             & 33.5                              & 37.1                           & 32.3                               & 31.2                                    & 27.6                               & 42.7                         & 43.4                             & 44.7                                  & \textbf{45.3}                    & 36.7                          & 37.0                              & 35.8                                   & 33.6                              \\
appeq/enteq/wc, sboisa & 38.2                         & 37.4                             & 37.4                                  & 37.6                             & 38.2                              & 42.0                           & 36.6                               & 34.3                                    & 29.1                               & 49.4                         & 48.9                             & 50.4                                  & \textbf{52.0}                    & 41.7                          & 42.1                              & 41.0                                   & 39.0                              \\
appeq/entov/wc, sboeq  & 25.7                         & 25.1                             & 24.8                                  & 25.3                             & 26.3                              & 26.4                           & 18.3                               & 16.9                                    & 18.0                               & 36.7                         & 34.1                             & 38.3                                  & \textbf{39.7}                    & 28.7                          & 28.6                              & 28.0                                   & 27.5                              \\
appeq/entov/wc, sboov  & 33.5                         & 32.7                             & 32.5                                  & 32.4                             & 33.5                              & 37.1                           & 32.3                               & 31.2                                    & 27.6                               & 42.7                         & 43.4                             & 44.7                                  & \textbf{45.3}                    & 36.7                          & 37.0                              & 35.8                                   & 33.6                              \\
appeq/entov/wc, sboisa & 38.2                         & 37.4                             & 37.4                                  & 37.6                             & 38.2                              & 42.0                           & 36.6                               & 34.3                                    & 29.1                               & 49.4                         & 48.9                             & 50.4                                  & \textbf{52.0}                    & 41.7                          & 42.1                              & 41.0                                   & 39.0                              \\ 
\hline\end{tabular}
\caption{Supplementary materials: Macro F-score all results}
\label{t:macro-f-score-results}

\end{sidewaystable*}
\end{document}